\documentclass[10pt,journal,compsoc]{IEEEtran}
\usepackage{etoolbox}
\usepackage{caption}
\usepackage{subcaption}
\usepackage{amsmath,amsfonts}
\usepackage{algorithmic}
\usepackage{array}
\usepackage{textcomp}
\usepackage{stfloats}
\usepackage{url}
\usepackage{verbatim}
\usepackage{graphicx}
\usepackage[colorlinks,linkcolor=red,anchorcolor=blue,citecolor=green]{hyperref}
\def\BibTeX{{\rm B\kern-.05em{\sc i\kern-.025em b}\kern-.08em
    T\kern-.1667em\lower.7ex\hbox{E}\kern-.125emX}}
\usepackage{balance}
\usepackage{color}
\usepackage{amssymb}
\usepackage{amsmath}  % 必须包含，用于 split 和 text
\usepackage{bm}
\usepackage{wrapfig}
\usepackage{bbm}
\usepackage{mathrsfs}
\usepackage{algorithm}
\usepackage{algorithmic}
\usepackage{multirow}
\usepackage{bm}
\usepackage{bbding}
\usepackage{pifont}
\usepackage[table]{xcolor}

\usepackage{colortbl}
\definecolor{Gray}{gray}{0.9}

\newcolumntype{I}{!{\vrule width 1pt}}
\usepackage[colorlinks,linkcolor=red,anchorcolor=blue,citecolor=green]{hyperref}
\usepackage[super]{nth}



\usepackage{arydshln}
\definecolor{mydarkblue}{rgb}{0,0.08,0.45}

\usepackage[utf8]{inputenc}

\usepackage{listings}

\definecolor{bblue}{RGB}{0,30,95}
\definecolor{rred}{RGB}{190,0,0}
\definecolor{mygray}{gray}{.9}
\definecolor{ggray}{RGB}{127,127,127}
\definecolor{sblue}{RGB}{0,173,206}
\definecolor{ppink}{RGB}{240,46,142}
\newcommand{\ie}{\textit{i}.\textit{e}.}
\newcommand{\eg}{\textit{e}.\textit{g}.}

\newcommand{\cf}{\textit{cf}.}
\newcolumntype{y}[1]{>{\raggedright\arraybackslash}p{#1pt}}
\newcolumntype{z}[1]{>{\raggedleft\arraybackslash}p{#1pt}}

\usepackage{tabularx}
\definecolor{patch}{RGB}{216,118,52}
\definecolor{frame}{RGB}{77,85,129}
\definecolor{proto}{RGB}{112,48,160}
\definecolor{temp}{RGB}{64,136,201}

\def\eg{\emph{e.g.}} 
\def\vs{\emph{vs.}} 
\def\ie{\emph{i.e.}} 
\def\cf{\emph{cf.}}

\newcommand{\our}{\textsc{SeeTok}}
\usepackage{multirow}
\usepackage{multicol}
\makeatletter
\newcommand{\thickhline}{%
    \noalign {\ifnum 0=`}\fi \hrule height 1pt
    \futurelet \reserved@a \@xhline
}
\makeatother

\usepackage[table]{xcolor}
\usepackage{arydshln} 
\definecolor{mygray}{gray}{.9}
\definecolor{bblue}{RGB}{0,30,95}
\definecolor{rred}{RGB}{190,0,0}
\definecolor{mygray}{gray}{.9}
\definecolor{myred}{RGB}{168,71,81}
\definecolor{ggray}{RGB}{127,127,127}
\definecolor{sblue}{RGB}{0,173,206}
\definecolor{ppink}{RGB}{240,46,142}
\definecolor{myblue}{RGB}{206,219,242}

\usepackage{bm}
\usepackage{bbding}
\usepackage{pifont}
\newcommand{\cmark}{\ding{51}}
\usepackage{booktabs}
\definecolor{darkergreen}{RGB}{21, 152, 56}
\definecolor{red2}{RGB}{252, 54, 65}

\newcommand\greenp[1]{\textcolor{darkergreen}{(#1)}}
\usepackage{tikz} 

\usepackage[normalem]{ulem} % 用于删除线效果

\usepackage{pifont}

\usepackage[table,xcdraw,dvipsnames]{xcolor}
\definecolor{citecolor}{HTML}{0071BC}
\definecolor{linkcolor}{HTML}{ED1C24}
\usepackage{microtype}

\usepackage{graphicx}

\usepackage{booktabs}  % for \toprule, \midrule, \bottomrule
\usepackage{tabularx}  % for adjustable-width columns
\usepackage[utf8]{inputenc}
\usepackage{array}     % for p{width} column type

\usepackage[textsize=tiny]{todonotes}
\newcommand{\colorboxinline}[1]{\textcolor{myblue}{\rule{0.8em}{0.8em}}}
\usepackage[utf8]{inputenc}
\usepackage{newunicodechar}
\newunicodechar{♫}{\ding{14}} % or another suitable symbol

\begin{document}
\title{See the Text: \\ From Tokenization to Visual Reading}

\author{Ling Xing, Rui Yan, Alex Jinpeng Wang,~\IEEEmembership{Member, IEEE}, Zechao Li,~\IEEEmembership{Senior Member, IEEE}, Jinhui Tang,~\IEEEmembership{Senior Member, IEEE}
\thanks{

\textit{L. Xing, R. Yan, and Z. Li are with the School of Computer Science and Engineering, Nanjing University of Science and Technology, Nanjing 210094, China. (E-mail: \{lingxing, ruiyan, zechao.li\}@njust.edu.cn.) }
        
\textit{A. J. Wang is with the School of Computer Science and Engineering, Central South University, Changsha 410083, China. (E-mail: jinpengwang@csu.edu.cn)}

\textit{J. Tang is with the College of Information Science and Technology and Artificial Intelligence, Nanjing Forestry University, Nanjing 210037, China. (E-mail: tangjh@njfu.edu.cn)}
}
}
\markboth{ }%
{Shell \MakeLowercase{\textit{et al.}}: A Sample Article Using IEEEtran.cls for IEEE Journals}

\IEEEtitleabstractindextext{
\begin{abstract}

%Humans read text visually. 
People see text.
Humans read by recognizing words as visual objects, including their shapes, layouts, and patterns, before connecting them to meaning, which enables us to handle typos, distorted fonts, and various scripts effectively.
Modern large language models (LLMs), however, rely on subword tokenization, fragmenting text into pieces from a fixed vocabulary. 
While effective for high-resource languages, this approach over-segments low-resource languages, yielding long, linguistically meaningless sequences and inflating computation.
In this work, we challenge this entrenched paradigm and move toward a vision-centric alternative. 
Our method, \our, renders text as images (visual-text) and leverages pretrained multimodal LLMs to interpret them, reusing strong OCR and text–vision alignment abilities learned from large-scale multimodal training.
Across three different language tasks, \our\ matches or surpasses subword tokenizers while requiring 4.43× fewer tokens and reducing FLOPs by 70.5\%, with additional gains in cross-lingual generalization, robustness to typographic noise, and linguistic hierarchy.
\our\ signals a shift from symbolic tokenization to human-like visual reading, and takes a step toward more natural and cognitively inspired language models.

\end{abstract}
}

\maketitle

\IEEEdisplaynontitleabstractindextext
\IEEEpeerreviewmaketitle

\section{Introduction}
\label{sec:intro}

\begin{quote}
	\it\small Huamn mnid deos not raed ervey lteter by istlef, but the wrod as a wlohes~\cite{rawlinson1976thesis}. \\
	\mbox{}\hfill -- Graham Rawlinson
	%%\vspace{-3pt}
\end{quote}
\IEEEPARstart{E}{ven} with internal letters scrambled, humans can reconstruct the intended words with remarkable ease. 
The striking phenomenon, commonly referred to as \emph{typoglycemia}~\cite{johnson2007transposed}, highlights the profound robustness of human reading.
Psychologists found that this ability is rooted in the Visual Word Form Area (VWFA), a brain region that identifies familiar words from visual word shapes~\cite{dehaene2011unique,mccandliss2003visual,wimmer2016visual}.
Scrambled words typically preserve their overall shape and salient letter features, which allows the VWFA to tolerate noisy inputs and recover the intended words~\cite{rayner2012psychology,agrawal2020compositional}.
By leveraging holistic visual patterns and morphological cues, humans not only read efficiently and maintain robustness against noisy text~\cite{wang2024fast}, but can also acquire multiple languages and writing systems with remarkable flexibility~\cite{cohen2002language,dehaene2010reading}.

In contrast, modern LLMs~\cite{bai2025qwen2,zhang2024tinyllama,li2025uni} follow a strikingly different path, leaning heavily on subword tokenization techniques, such as Byte-level BPE~\cite{wang2020neural}, which break text into discrete subword units from a fixed vocabulary, shaping a unique narrative of how machines process language.
While effective for high-resource languages like English, this approach discards the continuous visual and morphological cues inherent in written languages. 
This makes tokenization highly sensitive to typos and minor perturbations~\cite{chai2024dual}, which can significantly disrupt token sequences, with no ability to leverage visual similarity for correction.
In multilingual contexts, it forces a compromise between inadequate coverage for low-resource languages and impractically large vocabularies~\cite{rust2022language}.

\begin{figure*}[!t]
    \centering
    \includegraphics[width=1.0\textwidth]{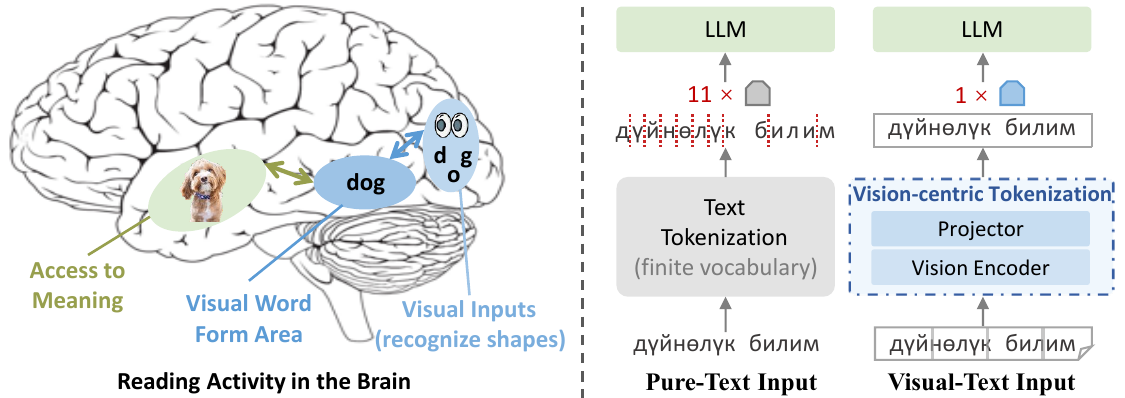}

    \caption{\small
  % \textbf{Breaking the vocabulary barrier.}
 \textbf{Left:}
Reading proceeds through a \textbf{visual–linguistic pathway}: the \textbf{visual} stream identifies letter shapes and patterns in the visual cortex and packages them into recognizable word forms via visual word form area; the \textbf{linguistic} stream in the left-hemisphere derives meaning.
\textbf{Right:} Subword tokenization tends to \emph{over-segment low-resource languages} due to insufficient vocabulary coverage, \eg, a 2-word Kyrgyz phrase (``\textit{world knowledge}") is split into 11 text tokens.
Vision-centric tokenization instead compresses the phrase into a single visual token by aggregating features from four adjacent image patches through the projector.
    }
    \label{fig:movtivation}
\vspace{-0.2cm}
\end{figure*}

We rethink the entrenched subword tokenization in LLMs and turn to a more \emph{human-like} approach. 
The human brain is highly plastic, leveraging a shared \textbf{visual–linguistic pathway} across languages to map word shapes onto meanings seamlessly, as shown in Figure~\ref{fig:movtivation} (left). 
Inspired by this mechanism, we introduce \our, a simple yet powerful vision-centric tokenization method for LLMs.
Specifically, \our\ first renders text into images and leverages the visual encoders of pretrained MLLMs (\eg, Qwen2.5-VL~\cite{bai2025qwen2}) to extract textual representations, which are then passed to the LLM backbone for deeper processing. 
Benefiting from large-scale vision-language pretraining, these visual encoders naturally exhibit strong OCR ability and robust text–vision alignment~\cite{yao2025efficient,lin2025parrot,liu2024ocrbench}, making them a promising alternative to conventional text tokenization.
To enhance instruction-following in the visual modality, we introduce vision-centric instruction tuning, where instruction texts are rendered as images (\ie, visual-text instructions) and the MLLM is adapted with lightweight LoRA~\cite{hu2022lora} layers. 
This simple yet effective procedure enables MLLMs to interpret visual-text instructions on par with pure-text ones, without costly training from scratch or architectural modifications.

We primarily evaluate our \our\ on the widely-used open-source models JanusPro~\cite{chen2025janus}, Qwen2.5-VL~\cite{bai2025qwen2} and Llava-next~\cite{liu2024llavanext}.
Across three representative natural language understanding tasks, \our\ achieves performance on par with text-tokenization baseline, while requiring \textbf{4.43×} fewer visual tokens and reducing FLOPs by \textbf{70.5\%}.
In multilingual translation covering 13 languages, \our\ further shows stronger cross-lingual transfer compared to the text-tokenization counterpart, achieving \textbf{86\%} lower fertility (\ie, fewer tokens per word) and a \textbf{+3.87} gain in COMET-22 scores.
Moreover, \our\ exhibits robustness to input perturbations (\cf~Sec.~\ref{sec:perturbation}), showing substantially smaller performance drops than the text-tokenization model across character-level, word-level, and visual-level attacks.
Beyond robustness, \our\ preserves intra-token structure lost in text tokenization, improving the modeling of character-level composition (\cf~Sec.~\ref{subwordcom}). 
This structural awareness enables better performance on fine-grained tasks such as character counting (\cf~Sec.~\ref{sec:ccount}) and word unscrambling (\cf~Sec.~\ref{sec:wordunscram}), which are challenging for text tokenization-based models.

Below, we summarize the advantages of our vision-centric tokenization, highlighting that representing text visually is a promising and valuable direction for future research.
\ding{182} \textbf{Efficiency.} 
Compared to text tokenization, our vision-centric tokenization significantly reduces token counts \emph{across 14 diverse languages} (\cf~Figure~\ref{fig:compression_ratio}), with even greater benefits for low-resource languages (\eg, 4.43× for English, 13.05× for Georgian).  
This advantage arises from its language-agnostic design, avoiding the inherent bias of text tokenization toward high-resource languages~\cite{truong2024revisiting}.
\ding{183} \textbf{Strong cross-lingual generalization.} Our vision-centric tokenization demonstrates robust cross-lingual generalization, achieving higher translation quality than text-tokenization counterpart for both high- and low-resource languages while avoiding excessive subword segmentation (\cf~Sec.~\ref{sec:multilingual}).
\ding{184} \textbf{Intra-token Structural Awareness.}
While text tokenization is often blind to the internal structure of tokens~\cite{chai2024tokenization}, our vision-centric paradigm preserves the compositional details of text.
This awareness yields inherent robustness to surface-level perturbations (\eg, typos and visual noise, \cf~Sec.~\ref{sec:perturbation}) and facilitates fine-grained structural reasoning tasks—including character counting and unscrambling (\cf~Sec.~\ref{sec:intratoken})—which are challenging for models that rely on discrete token IDs. 
\ding{185} \textbf{Input Flexibility.}
Representing text visually opens a new frontier for input customization.
Modulating rendering configurations (\eg, font size, resolution) allows for a dynamic trade-off between computational cost and performance. 
Simultaneously, stylistic variations (\eg, bolding, highlighting, italic) provide an intuitive way to emphasize key content.
This level of control is structurally impossible for traditional tokenization, which treats text as static, discrete IDs.

\section{Related Work}
\label{sec:related}
\subsection{Text Tokenization} 
Text tokenization~\cite{kenton2019bert,kudo-richardson-2018-sentencepiece, sennrich-etal-2016-neural} is the first step in natural language processing, segmenting the strings of text into smaller units.
Based on the granularity of segmentation, tokenization can be broadly classified into three types.
1) \emph{Character-level} tokenization treats each character or byte as an atomic token~\cite{xue2022byt5}. 
This design keeps the vocabulary small, but results in long input sequences that substantially increase memory and computation costs. 
Several strategies~\cite{yu2023megabyte,pagnoni2024byte} have been developed to mitigate this limitation.
2) \emph{Word-level} tokenization operates on entire lexical items, typically segmented by whitespace or language-specific heuristics~\cite{bengio2003neural}. 
They are efficient for frequent words, but face out-of-vocabulary (OOV) issues and demand huge vocabularies in multilingual settings, which inflate memory usage and make the softmax in the output layer computationally expensive.
3) \emph{Subword-level} tokenization, such as BPE~\cite{sennrich2016neural}, WordPiece~\cite{devlin2019bert}, and Unigram~\cite{kudo2018subword}, segment words into subword units and are now widely used. 
They balance vocabulary size and coverage while mitigating OOV issues, but break morphological boundaries and are sensitive to surface noise~\cite{rust2022language}.
In multilingual contexts, the \emph{fixed} vocabulary is \emph{primarily allocated to high-resource languages}, leaving low-resource languages with limited coverage.  
Consequently, words in \emph{low-resource languages are over-segmented}, sometimes almost at \emph{character-level}, leading to significantly longer token length.
In this work, we explore a vision-centric tokenization route that treats raw text as images.
This method promotes multilingual fairness, achieving \emph{low token fertility even for low-resource languages}.

\subsection{Enhancing Text Understanding through Visual Inputs}
Textual information often appears as part of visual data in real-world images, such as in scene text, documents, and charts~\cite{li2025monkeyocr,huangmini,lin2024parrot,DuCSJJ25}. 
Modeling such visual text has been a long-standing problem in computer vision and multimodal learning~\cite{shi2016end,wang2025wilddoc,wang2023enhancing}.
Early approaches rely on optical character recognition (OCR) to extract symbolic text~\cite{huang2022layoutlmv3,lyu2018mask,shi2018aster}, followed by language modeling.
More recently, OCR-free large multimodal models~\cite{liu2026textmonkey,fuocrbench,gao2024improving,lotz2023text,zhuang2025math,zhang2024cross,rust2022language,lee2023pix2struct,liao2022real} have demonstrated that text understanding can be achieved directly from raw images, bypassing explicit transcription.
Visual text representations have also been applied to other domains such as machine translation~\cite{salesky2021robust} and long-context compression~\cite{wang2024leveraging,xing2025vision,cheng2025glyph}, where visual text provides compact embeddings that support efficient processing.
While these approaches indicate the potential of visual inputs for language understanding, they generally treat visual text as an auxiliary modality and still rely on subword tokenization as the primary interface~\cite{wang2024leveraging,cheng2025glyph,liu2026textmonkey}.
In contrast, our method bypasses subword tokenization entirely and moves toward a vision-centric alternative.

\section{Methodology}
\label{sec::method}

\our\ proposes a novel approach in which text is not fed as discrete tokens but rendered into images, enabling the model to perceive and process textual content visually (visual-text).

\subsection{Overall Pipeline}
\begin{figure}[!t]
  \centering
  % 在双栏排版中，\columnwidth 恰好等于单栏的宽度
  \includegraphics[width=0.8\columnwidth]{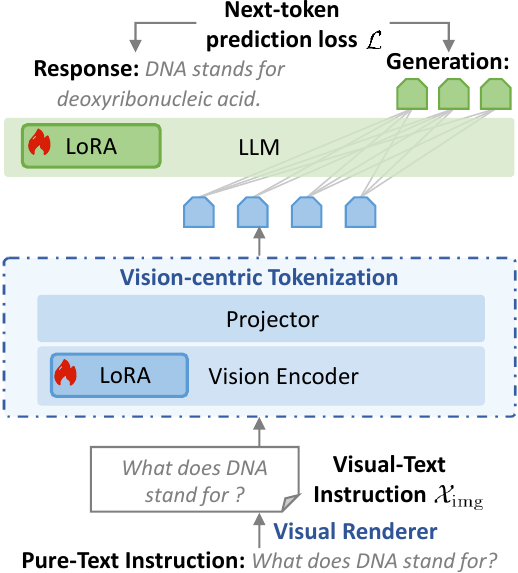}
  
  \vspace{-5pt} % 适当缩小图片与 Caption 之间的间距
  \caption{\small
    \textbf{Overview of \our.}
    Text is rendered into an image, processed by the vision-centric tokenization, and fed to the LLM.
    The integration of LoRA layers further aligns the model with complex visual-text instructions, enhancing its capacity for instruction-driven reasoning.}
  \label{fig:overview}
  
  \vspace{-10pt} % 适当缩小 Caption 与下方正文之间的间距，有助于节省空间
\end{figure}

Figure~\ref{fig:overview} illustrates the overall pipeline of \our. 
Given an input text sequence, we first apply a \textbf{visual renderer} that transforms the raw string into a rendered text image.
The image is then processed by the \textbf{vision-centric tokenization} (\ie, vision encoder and MLP projector from MLLMs), which substitutes for standard text tokenization and empowers the LLM to perceive text directly in visual form rather than as discrete tokens. 
The LLM subsequently consumes these encoded visual features to perform downstream reasoning and generation.
We primarily base our study on widely used Qwen2.5-VL 3B/7B~\cite{bai2025qwen2}, Llava-next~\cite{liu2024llavanext}, and JanusPro~\cite{chen2025janus}.

Although modern MLLMs possess strong OCR and vision–language alignment~\cite{yao2025efficient,lin2025parrot}, they are rarely exposed to \emph{visual-text instructions} (\ie, instructions presented as rendered images) during pretraining. 
This results in a distribution gap, causing weaker visual-text instruction-following ability compared to pure-text instructions.
To close the gap, we integrate LoRA adapters~\cite{hu2022lora} into both the vision encoder and the LLM.
LoRA adapters improve fine-grained text perception on the vision encoder side and align instruction-following on the LLM side, enabling \our\ to handle visual-text prompts effectively with negligible training overhead compared to pretraining from scratch, while incurring no additional inference parameters.

\subsection{Visual Text Tokenization}
\noindent\textbf{Visual Renderer.} 
The core component of \our\ is a visual renderer that transforms raw textual data into RGB images $\mathcal{X}_\text{img}\!=\!\{x_m \!\in\! \mathbb{R}^{H\times W \times C}\}_{m=1}^{M}$, where $M$ denotes the number of rendered text images and can be dynamically adjusted based on the length of the input text. 

If vision encoder supports variable resolutions (\eg, the 3B and 7B versions of Qwen2.5-VL~\cite{bai2025qwen2}), both the image height $H$ and width $W$ can be scaled to match the text length. 
In our training setup, we configure $M=1$ with height $H=14$, width $W=3584$, and $C=3$ channels, corresponding to an image of resolution $224 \times 224$. 
Text is rendered using the \textit{Google Noto Sans} typeface with a font size of 7px.  

\noindent\textbf{Vision-centric Tokenization.}
The visual-text is first processed by the vision encoder to extract patch-level features. 
On average, a $14\times14$ image patch encodes roughly 1.1 Qwen tokens in English, highlighting the compactness of the visual representation. 
A two-layer MLP projector then aggregates four neighboring patches and projects them into a dimension aligned with the text embeddings of the LLM, reducing the token sequence length by 4$\times$.
Together, the vision encoder and projector function as a \emph{``visual'' text tokenization}, providing an efficient and effective substitute for standard text tokenization.
In contrast to \emph{subword vocabulary biased toward high-resource languages}, patch-based segmentation ensures that diverse languages are encoded fairly \textbf{without requiring vocabulary enlargement}. 
This design yields substantial efficiency gains compared to text tokenization, reducing fertility (\ie, the average token count per word) by \textbf{86\%} on average across 13 languages, including both high- and low-resource languages (\cf~Sec.~\ref{sec:multilingual}).

% \subsection{Improving Instruction Following with Text-Image Inputs}
\subsection{Vision-centric Finetuning}
Pretrained MLLMs demonstrate strong OCR capabilities and excel at recognizing textual content within images~\cite{yao2025efficient,lin2025parrot,liu2024ocrbench,qu2026spatio}.  
However, when instructions are provided as visual-text instead of pure-text, model performance drops significantly (\cf~Table~\ref{tab:table1}).  
This indicates that, although the model can accurately read the text, it struggles to interpret it as an instruction and perform reasoning accordingly.  
This gap may be because the MLLMs are rarely exposed to visual-text instructions during pretraining, and thus fail to associate visualized text with the same instruction-following semantics as conventional text tokens.

To address this limitation, we perform instruction tuning using LoRA layers~\cite{hu2022lora} applied to both the vision encoder and the LLM.  
During tuning, instructions are rendered as text images, while target answers remain in textual form to compute the next-token prediction loss.
Formally, given an instruction $I$ rendered as images $\mathcal{X}_\text{img}$ and a target response sequence $\mathbf{y} = (y_1,\dots,y_T)$, we optimize the standard autoregressive generation loss:  
\begin{equation}
\mathcal{L} = - \sum_{t=1}^{T} \log P \big( y_t \mid y_{<t}, \mathcal{X}_\text{img} \big),
\end{equation}
where $\mathcal{X}_\text{img}$ is encoded via the vision encoder and MLP to serve as the instruction signal, and the decoder LLM generates the answer token by token.  
This training explicitly encourages the model to interpret visualized instructions correctly and generate responses that align with textual answers.
By leveraging pretrained MLLMs, \our\ realizes a vision-centric tokenization in a lightweight and efficient manner, \textbf{eliminating the need to train from scratch}.
Crucially, our experiments indicate that \emph{keeping the projector frozen} is essential for stable performance during instruction tuning (\cf~Table~\ref{tab:lora}), as it preserves the robust cross-modal alignment learned from large-scale pretraining.
Beyond bridging the gap between visual-text and pure-text prompts, \our\ enables efficient multilingual representation, strong compositional reasoning, and robustness to typographical or visual perturbations (\cf~Sec.~\ref{sec:perturbation}), while preserving the original architecture and vocabulary. 
These properties underscore the advantages of our approach and motivate further exploration.

\subsection{Why not Vision-centric Detokenization?}
\label{app:vision_centric}
The primary computational bottleneck in current LLMs arises from their large parameter sizes and the quadratic complexity of Transformer self-attention with respect to \textbf{input length}. 
\textit{Text detokenization at the output stage remains computationally negligible}.
We therefore focus on reducing input token length via our vision-centric tokenization while retaining the pretrained text detokenization.
This asymmetric design offers several strategic advantages:
\textbf{(i) Knowledge Preservation}: Reusing the standard text detokenizer allows the model to effectively reuse pretrained linguistic knowledge and prevents catastrophic forgetting.
\textbf{(ii) Architectural Seamlessness}: Reusing text detokenizer preserves the existing MLLM architecture, thereby allowing seamless application of our method to more vision-encoder-based MLLMs.
\textbf{(iii) Efficiency}: By generating discrete text tokens, we bypass the need for external OCR systems and avoid the potential sources of error and heavy overhead associated with high-fidelity image synthesis.

In contrast, developing a vision-centric output paradigm introduces non-trivial challenges: achieving high orthographic fidelity, managing computational costs of pixel-level generation, and implementing adaptive layouts to avoid redundant padding (blank space). 
Such generative complexities constitute a distinct research trajectory and lie beyond the scope of this work.

\section{Experiment}
\label{sec::experi}

\subsection{Experimental Setup}
\label{sec:exp:setup}

\begin{table*}[t!]

\captionsetup{font=small}
\caption{ \small{Our vision tokenization-based \our\ significantly enhances Qwen2.5-VL 3B with visual-text input on diversity language understanding tasks. 
On average across multiple types of language tasks, \our\ matches the performance of the text-tokenization baseline Qwen2.5-VL 3B with pure-text input.}
}
% \vspace{-0.3cm}
\centering
\small
\resizebox{1.\textwidth}{!}{
\renewcommand\arraystretch{1.4}
\begin{tabular}{cccccccc}
\thickhline

\textbf{Models}  & \textbf{Text Source}  & \textbf{TriviaQA}   & \textbf{NQ}  & \textbf{PopQA} & \textbf{MMLU}  & \textbf{SST5} & \textbf{Avg.} \\
\hline

 Qwen2.5-VL 3B & Pure-Text & 41.92 &29.31	&24.64 &	61.91	&28.80&	37.32 \\

 \arrayrulecolor{gray}\hdashline\arrayrulecolor{black}

Qwen2.5-VL 3B & Visual-Text & 37.55 	&21.13 &	20.16	&32.31	&25.21	&27.27 \\

 + \our & Visual-Text & \textbf{43.53\greenp{5.98$\uparrow$}}&	24.14\textbf{\greenp{3.01$\uparrow$}}&	24.26\textbf{\greenp{4.10$\uparrow$}}&
 52.52\textbf{\greenp{20.21$\uparrow$}}
 &\textbf{44.40\greenp{19.19$\uparrow$}}&	
 \textbf{37.77}\textbf{\greenp{10.50$\uparrow$}} \\
\bottomrule
\end{tabular}}
\vspace{-0.2cm}
\label{tab:table1}
\end{table*}

\noindent\textbf{Instruction-tuning Dataset Details.} We employ OpenHermes 2.5~\cite{OpenHermes2.5} as the instruction-tuning corpus by default, providing a larger-scale and high-quality collection of diverse instruction–chat samples.
Due to resource limitations, we exclude excessively long samples to prevent out-of-memory issues, resulting in a filtered corpus of 658k instances.

\noindent\textbf{Multilingual Dataset Details.}  
We evaluate the multilingual proficiency of \our\ across 13 diverse languages, stratified into five high-resource and eight low-resource groups. 
The high-resource set comprises German (de), Czech (cs), Icelandic (is), Chinese (zh), and Russian (ru). 
To ensure a comprehensive assessment, the low-resource set is curated to encompass a broad spectrum of typological features and linguistic families, including: Turkic (Kyrgyz, ky; Uzbek, uz), South Caucasian (Georgian, ka), Baltic (Lithuanian, lt; Latvian, lv), Slavic (Bulgarian, bg; Macedonian, mk), and Austronesian (Malagasy, mg).
For high-resource languages, we finetune \our\ on ALMA~\cite{xuparadigm}, which collects human-written test datasets from WMT’17 to WMT’20, plus the dev and test sets from Flores-200~\cite{costa2022no}, resulting in a total of 58K training examples across all languages. 
For low-resource languages, in line with X-ALMA~\cite{citation-0}, we use the Flores-200 dev set~\cite{costa2022no} as training data to ensure high data quality. 

\noindent\textbf{Downstream Evaluation.} 
We evaluate \our\ against traditional text-tokenization counterpart, focusing on four key axes: 
general-purpose proficiency, cross-lingual versatility, robustness to input perturbations, and intra-token structural fidelity.
(1) \textbf{General Knowledge and Reasoning}: Evaluation on five representative downstream benchmarks to assess the core reasoning and knowledge capabilities (\cf~Sec.~\ref{sec:nlp}).
(2) \textbf{Multilingual Versatility and Efficiency}: Assessment of multilingual transfer across 13 languages, comprising 5 high-resource and 8 low-resource languages (\cf~Sec.~\ref {sec:multilingual}).
(3) \textbf{Robustness to Input Perturbations}: Evaluation of model resilience against \textit{character-, word-, and visual-level noise} to measure its reliability under textual corruptions (\cf~Sec.~\ref{sec:perturbation}).
(4) \textbf{Intra-token Structural Fidelity}: To quantify fine-grained lexical perception, we utilize \textit{subword composition, character counting, and word unscrambling} tasks (\cf~Sec.~\ref{sec:intratoken}).

\subsection{Implementation Details} 
To demonstrate the generalizability of \our, we conduct an extensive evaluation across a broad spectrum of representative MLLMs, including Qwen2.5-VL 3B~\cite{bai2025qwen2}, Qwen2.5-VL 7B~\cite{bai2025qwen2}, and LLaVA-NeXT 8B~\cite{liu2024llavanext}.
Crucially, we additionally extend our evaluation to JanusPro~\cite{chen2025janus}, a unified model that integrates multimodal understanding and generation into a cohesive architecture. 
Evaluating \our\ across these heterogeneous systems demonstrates its robust performance regardless of the underlying model architecture.
To reduce computational overhead, we employ DeepSpeed with ZeRO stage-2~\cite{rasley2020deepspeed} and float16 precision.
We employ LoRA~\cite{hu2022lora} for instruction tuning, injecting low-rank adapters with rank $r=8$, scaling factor $\alpha=32$, and 10\% dropout.
All bias parameters are kept frozen during training.
For optimization, we use the AdamW optimizer~\cite{loshchilov2017decoupled} at a peak learning rate of $2\times10^{-5}$ and a weight decay of 0.1. 
The schedule begins with a linear warm-up from $1\times10^{-7}$ over the first 1000 steps, after which the learning rate decays exponentially to zero. 
Global gradient clipping with a threshold of 1.0 is employed to maintain training stability.
For validation on JanusPro~\cite{chen2025janus}, which requires $384 \times 384$ input images, we configure the input with $M=1$, height $H=16$, width $W=9216$, and $C=3$ channels. 
This configuration aligns with the spatial dimensions of a $384 \times 384$ square image, ensuring seamless compatibility with the vision encoder of JanusPro~\cite{chen2025janus}.

For language understanding tasks, we evaluate on MMLU~\cite{hendrycksmeasuring} using a zero-shot setup, and on SST5~\cite{socher2013recursive} with 5-shot sampling. 
For question answering tasks (TriviaQA~\cite{joshi2017triviaqa}, NQ~\cite{kwiatkowski2019natural}, and PopQA~\cite{mallen2023not}), we employ Contriever~\cite{izacardunsupervised} to retrieve the top-$k$ relevant passages from Wikipedia, following the CEPE protocol~\cite{yen2024long}. 
We prioritize providing the most relevant passages to the decoder to improve performance. 

\subsection{Performance on Comprehensive Benchmarks}
\label{sec:nlp}
To assess the effectiveness of our \our, we evaluate on multiple representative natural language understanding tasks, spanning open-domain question answering (TriviaQA~\cite{joshi2017triviaqa}, NQ~\cite{kwiatkowski2019natural}, and PopQA~\cite{mallen2023not}), general knowledge reasoning (MMLU~\cite{hendrycksmeasuring}, a massive multitask benchmark spanning 57 subjects that probes expert-level cognitive depth), and sentiment classification (SST5~\cite{socher2013recursive}).
We report Exact Match (EM) for QA and accuracy for MMLU and SST5.

\begin{table}[t] % [!t] 将表格置于本栏顶部，是双栏论文最标准的位置
  \centering
  \captionsetup{font=small}
  \caption{
    Evaluating efficiency between standard text tokenization (\ie, pure-text input) and vision tokenization(\ie, visual-text input) on the TriviaQA dataset~\cite{joshi2017triviaqa} based on \our. Compression ratio $\Delta$ is the ratio of the text-token count to the number of visual-text tokens.}
  \label{tab:efficiency}
  
  % \vspace{-5pt} % 缩小标题与表格之间的间隙
  
  % 使用 1.0\columnwidth 确保表格填满单栏宽度
  \renewcommand\arraystretch{1.2}
  \resizebox{0.9\columnwidth}{!}{
    
    \begin{tabular}{cccc}
      \toprule
      \textbf{Text Source} & \textbf{$\Delta$} & \textbf{Latency (s)} & \textbf{TFLOPs} \\
      \hline
      Pure-Text & - & 5.02 & 3.12 \\
      Visual-Text & 4.43 & \textbf{3.34} & \textbf{0.92} \\
      \hline
    \end{tabular}
  }
  \vspace{-0.2cm}
  % \vspace{-10pt} % 缩小表格与下方正文之间的间隙，节省空间
\end{table}
\subsubsection{Effectiveness}
As shown in Table~\ref{tab:table1}, \our\ (vision-centric tokenization over visual-text inputs) \textbf{matches or even surpasses} the text-tokenization counterpart (Qwen2.5-VL 3B), averaging 37.77 compared to 37.32 across five datasets.
Notably, \our\ outperforms the text-tokenization baseline on TriviaQA (+1.61) and SST5 (+15.60).
These two tasks rely heavily on surface-form cues such as spelling, capitalization, and negation.
Subword text tokenization often fragments or obscures such information, particularly for rare words and entities~\cite{tanaka2021visualmrc,truong2024revisiting}. 
In contrast, the vision-centric tokenization preserves character-level fidelity, enabling the model to capture these signals more faithfully.
MMLU~\cite{hendrycksmeasuring} is a knowledge-intensive benchmark spanning multiple domains, formulas, and logical reasoning, which relies more heavily on world knowledge learned from large-scale textual pretraining.
Since the vision pathway has not been exposed to comparable amounts of such data, a performance gap remains.
Similar pretraining conducted on visual-text could potentially further narrow this gap, a trend already reflected in the scaling experiments (see Sec.~\ref{app:scaling}).
Furthermore, our vision-centric paradigm provides a favorable performance-efficiency trade-off, achieving a \textbf{4.43$\times$ reduction} in sequence length compared to the text-tokenization baseline (detailed in Sec.~\ref{sec:effi}). 
This substantial gain in token-level efficiency directly translates to lower computational overhead while maintaining competitive performance across most tasks.
\begin{table}[t]
\centering
\small
\caption{ Comparison of FLOPs and memory usage between our \our\ and the text-tokenization based QwenVL 2.5 3B on long sequences (50k and 74k text tokens).}
\label{tab:app:memory_flops}
\renewcommand{\arraystretch}{1.2}
\begin{tabular}{cccc}
\toprule
\textbf{Model} & \textbf{Text Token Num} & \textbf{TFLOPs} & \textbf{Memory} \\
\midrule
\our & 50k & 129.55 & 15.6 GB \\
\our & 74k & 183.97 & 23.6 GB \\
Qwen2.5-VL 3B & 50k & 308.59 & 23.6 GB \\
\bottomrule
\end{tabular}
\vspace{-0.2cm}
\end{table}
\subsubsection{Efficiency} 
\label{sec:effi}
Vision-centric tokenization provides substantial efficiency benefits.
We quantify efficiency on TriviaQA~\cite{joshi2017triviaqa}, comparing two tokenization schemes: standard text tokenization and vision-centric tokenization. 
Both models are based on \our.
We report the compression ratio $\Delta$ defined as the dataset-level average text tokens divided by the average visual-text tokens, along with FLOPs and end-to-end latency (in seconds). 
As summarized in Table~\ref{tab:efficiency}, \our\ with visual-text input achieves \textbf{4.43× reduction in token length}, along with \textbf{70.5\% lower FLOPs} and \textbf{33.5\% faster latency} compared to the model with pure text input, while \textbf{maintaining comparable performance}.
These efficiency gains make vision-centric tokenization particularly attractive for resource-constrained environments, where reducing inference cost is critical.
Latency measures the total wall-clock time from input reception to the generation of 64 output tokens.

\noindent\textbf{Efficiency in Long Sequence Scenarios.}
To further evaluate the practical deployment potential of \our\ in long sequence scenarios, we conduct a rigorous analysis of its TFLOPs and memory consumption against the standard text-tokenized baseline (Qwen2.5-VL 3B) with long input.
As illustrated in Table~\ref{tab:app:memory_flops}, at a sequence length of 50k text tokens, \our\ reduces the computational overhead by 58.0\% (from 308.59 to 129.55 TFLOPs) and achieves a \textbf{33.9\% reduction} in peak GPU memory usage (from 23.6 GB to 15.6 GB). 
Within a strict 23.6 GB memory constraint, the text-tokenization baseline is limited to a 50k-token sequence, whereas \our\ successfully scales to 74k text tokens—representing a \textbf{48\% increase} in supported context length.
Remarkably, even when processing this extended 74k sequence, the total TFLOPs of \our\ (183.97) remains significantly lower than that of the baseline at only 50k tokens (308.59). 
These results underscore that \our\ effectively alleviates the memory-compute bottleneck, facilitating much broader contextual perception within identical hardware constraints.

\begin{table*}[t!]
%\vspace{0.1cm}
\captionsetup{font=small}
\caption{ \small{Translation performance from five high-resource languages to English. Fertility (FET) measures the average number of tokens used to represent a single word. COMET-22 score (COM) evaluates overall translation quality.
${\dagger}$ denotes the same LoRA setup as our \our, with pure-text training input.
}
}
% \vspace{-0.3cm}
\centering
\small
\resizebox{1.\textwidth}{!}{
\renewcommand\arraystretch{1.5}
\begin{tabular}{cccccccccccccc}
\thickhline

\multirow{2}{*}{\textbf{Models}} &\multirow{2}{*}{\textbf{Text Source}}
& \multicolumn{2}{c}{\textbf{de}}   & \multicolumn{2}{c}{\textbf{cs}}   & 
\multicolumn{2}{c}{\textbf{is}}   & 
\multicolumn{2}{c}{\textbf{zh}}   
& \multicolumn{2}{c}{\textbf{ru}}   &  \multicolumn{2}{c}{\textbf{Avg.}}   \\

\addlinespace[-0.5ex]
\cmidrule(lr){3-4} \cmidrule(lr){5-6} 
\cmidrule(lr){7-8}  \cmidrule(lr){9-10}  
\cmidrule(lr){11-12} \cmidrule(lr){13-14} 
\addlinespace[-0.5ex]

& & COM$\uparrow$  & FET$\downarrow$ &  COM$\uparrow$  & FET$\downarrow$ &
COM$\uparrow$  & FET$\downarrow$ &  COM$\uparrow$  & FET$\downarrow$ &
COM$\uparrow$  & FET$\downarrow$ &  COM$\uparrow$  & FET$\downarrow$ \\

\hline 

 Qwen2.5-VL 3B & Pure-Text & 67.25 &1.89 
 & 62.02& 2.81&	
 53.63&	2.71&
 57.51&  1.09&	
 63.16&	2.53
 &60.71  &2.21\\

{ Qwen2.5-VL 3B$^{\dagger}$}  & {Pure-Text} & {\textbf{67.88}} &{1.89 }
 & {62.05} &{ 2.81}
 & {53.89}&	{2.71
} & {58.12}& { 1.09}
 & {65.33} &{2.53
} &{61.45}  &{2.21
} \\

 \arrayrulecolor{gray}\hdashline\arrayrulecolor{black}

Qwen2.5-VL 3B & Visual-Text
&47.49&\textbf{0.42}
&41.02&\textbf{0.38}
&34.37&\textbf{0.37}
&46.77&\textbf{0.21}
&46.44&\textbf{0.49}
&33.72&\textbf{0.37}\\

+ \our & Visual-Text
 &65.63& \textbf{0.42}
 &\textbf{64.89}& \textbf{0.38}
 &\textbf{54.97}&\textbf{0.37}
 &\textbf{68.94}&\textbf{0.21}
 &\textbf{71.42}&\textbf{0.49}
 &\textbf{65.17}&\textbf{0.37}\\

\hline
\end{tabular}}
\vspace{-0.2cm}
\label{tab:high}
\end{table*}

\begin{figure}[!t]
    \centering
    \vspace{-0.3cm}
    \includegraphics[width=1.\columnwidth]{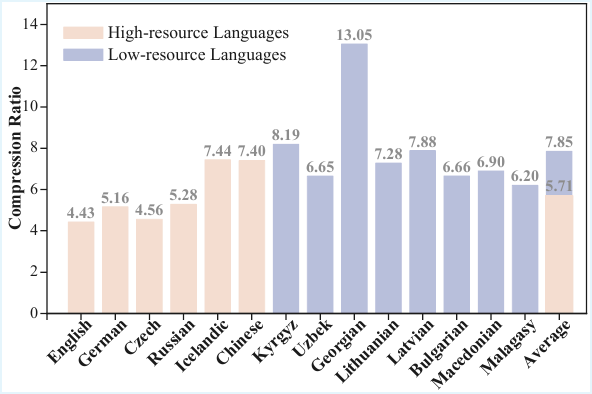} 
    \caption{\small \textbf{Compression Ratio Comparison.} 
    We calculate the ratio on Flores-200 test set~\cite{costa2022no} as the average length of sequences tokenized by standard text tokenization (Qwen2.5-VL 3B~\cite{qwen}) divided by the average length of the same sequences tokenized by our vision-centric tokenization (\our).
    Compared with text tokenization, our vision-centric tokenization achieves a compression ratio of \textbf{5.71$\times$} in high-resource languages and \textbf{7.85$\times$} in low-resource languages, significantly reducing token length.}
    \label{fig:compression_ratio}
    \vspace{-0.2cm}
\end{figure}

\subsection{Multilingual Translation Evaluation}
\label{sec:multilingual}
To evaluate the multilingual capabilities of our approach, we test the translation performance across multiple languages, divided into two groups:
\textbf{i) Five High-resource Languages:} de (German), cs (Czech), is (Icelandic), zh (Chinese), ru (Russian).
\textbf{ii) Eight Low-resource Languages:} ky (Kyrgyz), uz (Uzbek), ka (Georgian), lt (Lithuanian), lv (Latvian), bg (Bulgarian), mk (Macedonian), mg (Malagasy). 
We report the \textbf{COMET-22 score (COM)} for the translation from each of these languages to English~\cite{rei2022comet}, as suggested by \cite{freitag2023results}. 
A higher COM indicates better translation quality in terms of fluency and correctness.
We also calculate \textbf{Fertility (FET)}, a metric for assessing tokenization performance~\cite{rust2021good}, defined as the average number of tokens per word.
For word segmentation, we use Jieba for zh and whitespace splitting for other languages~\cite{ali2024tokenizer}.

% \begin{table}[t]
% %\vspace{0.1cm}
% \captionsetup{font=small}
% \caption{ \small Visualization of visual attack.
% }
% \centering
% \small
% \resizebox{0.3\textwidth}{!}{
% % \setlength\tabcolsep{8pt}
% \renewcommand\arraystretch{1.}
% \begin{tabular}{cc}
% \thickhline
% \textbf{Input Text}  & \textbf{Visual Attack}  \\
% \hline 
% a &	 â \\
% b &	 ḃ \\
% c &	 ĉ \\
% d &	 ḑ\\
% l&	 ᶅ\\
% H &	 Ĥ\\
% % V &	 Ṽ\\
% \hline
% \end{tabular}}

% \label{tab:app:visualnoise}
% \end{table}

\begin{table}[t]
\centering
\caption{\small Visualization of visual attack~\cite{eger2019text}.}
\label{tab:app:visualnoise}
\small
\renewcommand{\arraystretch}{1.} % 增加行高，匹配图片风格
\setlength{\tabcolsep}{20pt}      % 增加列间距，使其在单栏中更美观

\begin{tabular}{cc}
\toprule
\textbf{Input Text} & \textbf{Visual Attack} \\
\midrule
a & \^{a} \\
b & \.{b} \\
c & \^{c} \\
d & \c{d} \\
l & \c{l} \\
H & \^{H} \\
V & \~{V} \\
\bottomrule
\end{tabular}
\end{table}
\subsubsection{High-resource Languages}
Since the model Qwen2.5-VL 3B~\cite{bai2025qwen2} has not encountered multilingual visual-text instructions, we finetune it on ALMA~\cite{xuparadigm}, a small but high-quality bilingual corpus, to enable effective multilingual instruction following in the visual-text form.
To ensure fair comparison, Qwen2.5-VL 3B is also finetuned on the same dataset ALMA~\cite{xuparadigm} with pure-text input, denoted as Qwen2.5-VL 3B$^{\dagger}$.
Further training details are provided in Sec.~\ref{sec:exp:setup}. 
Following ALMA~\cite{xuparadigm}, we test on WMT22 test data~\cite{freitag2022results}, except for Icelandic (is), which is tested on WMT21~\cite{freitag2021results}.
Table~\ref{tab:high} illustrates that \our\ enhances the performance of Qwen2.5-VL with visual-text input, achieving an average COM improvement of \textbf{+31.45} across five languages.
Notably, \our\ outperforms the text-tokenization baseline Qwen2.5-VL 3B$^{\dagger}$ (65.17 vs. 61.45). 
This performance is achieved with substantially lower Fertility (FET); as shown in Table~\ref{tab:high}, \our\ requires only 0.37 tokens per word on average, a significant reduction from the 2.21 tokens used by standard text tokenization. 
This indicates that vision-centric tokenization \textbf{represents multilingual text more compactly while preserving translation quality}.
The performance advantage is particularly evident in non-Latin languages such as Chinese (zh) and Russian (ru), where \our\ achieves much higher COMET scores. 
This suggests that the vision-centric tokenization \emph{offers a stronger advantage for languages that differ more from English in terms of grammar and morphology}. 
For languages like German (de) and Czech (cs), the performance of \our\ is comparable to the Qwen2.5-VL 3B with pure text input, likely due to their syntactic similarities with English.

\begin{figure*}[!t]
    \centering
 
    \includegraphics[width=1.0\textwidth]{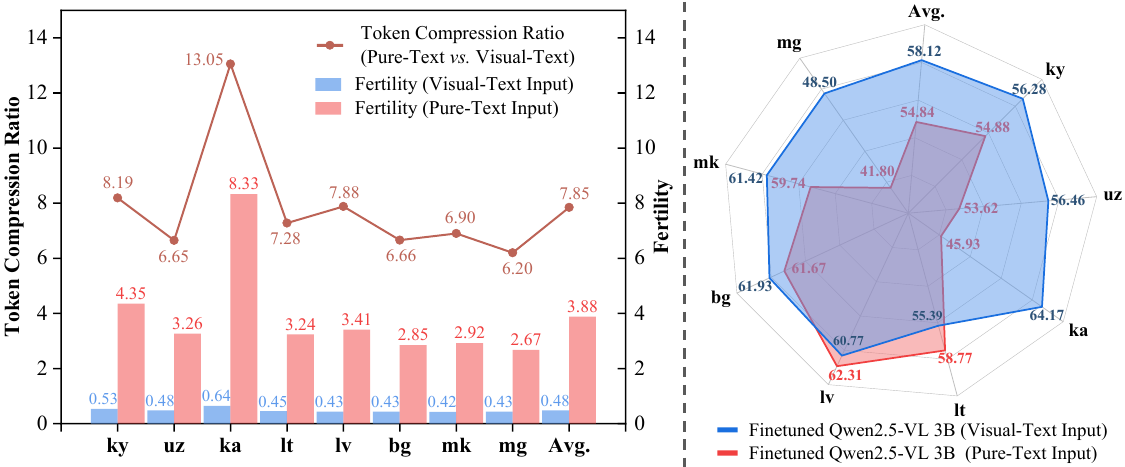} 
    
\vspace{-8pt}
    \caption{\small{\textbf{Left:} Fertility and token compression ratio across low-resource languages, comparing text and vision-centric tokenization. \textbf{Right:} COMET-22 scores on FLORES for translations from low-resource languages into English, comparing Qwen2.5-VL 3B trained with visual-text input and with pure-text input.}}
    \label{fig:lowsource}
\vspace{-0.2cm}
\end{figure*}

\begin{figure*}[!t]
    \centering
    \includegraphics[width=1.0\textwidth]{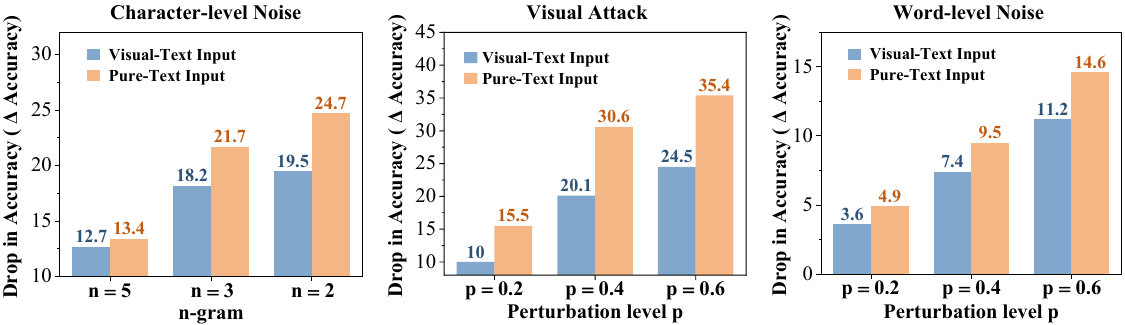} 
    \vspace{-0.2cm}

    \caption{\small \textbf{Accuracy drop} on MMLU~\cite{hendrycksmeasuring} under different orthographic perturbations (character-, visual-, and word-level noise). The vision tokenization–based model (blue) shows \textbf{markedly smaller declines} than the text-tokenization counterpart (orange), demonstrating stronger robustness to surface noise. }
    \label{fig:noise}
\vspace{-0.2cm}
\end{figure*}

\subsubsection{Low-resource Languages}
Since it is unclear whether Qwen2.5-VL 3B~\cite{bai2025qwen2} has seen these low-resource languages during pretraining, we finetune it using two methods: (i) pure text finetuning and (ii) visual-text finetuning. 
The training parallel data is provided by X-ALMA~\cite{citation-0}, and we use the FLORES test set~\cite{costa2022no} for evaluation.
As shown in Figure~\ref{fig:lowsource} (left), the average fertility (FET)~\cite{rust2021good} for these eight low-resource languages is 3.88 under standard subword tokenization (\ie, pure-text input). 
This means that, on average, just a single word is split into 3.88 fragments, resulting in excessive sequence lengths that hinder effective cross-lingual learning.
This is especially severe for Georgian (ka), where the FET peaks at 8.33.
In contrast, \our\ (\ie, visual-text input) maintains a remarkably low and consistent FET of 0.48, yielding an average \textbf{7.85$\times$ token compression ratio}. 
This drastic reduction in sequence length significantly alleviates the learning burden. 
Consequently, as shown in the radar chart (Figure~\ref{fig:lowsource}, right), \our\ achieves a higher average COMET score (58.12) compared to pure-text finetuning (54.84). 
These results suggest that by bypassing vocabulary-related biases, our vision-centric tokenization provides a more robust and efficient representation for languages that are poorly represented in traditional subword vocabularies.

\begin{table*}[h]
\centering
\small
\caption{\small Examples of text corruption with character- and word-level noise.
We calculate the similarity scores between the original text and the corrupted text computed by text tokenization and vision tokenization.
\textbf{Typoglycemia} refers to the phenomenon where words remain readable even when their \textbf{internal letters are scrambled}, as long as the first and last letters stay in place.
\textcolor{red}{Red} indicates letters whose order has been changed, \textcolor{blue}{blue} indicates letters that have been added or deleted, and \textcolor{green}{green} indicates letters that have been replaced with another character.}

% 修改列定义：将第三列设为 X，让其自动填满剩余空间；修正第四列 c 的语法
\begin{tabularx}{\textwidth}{X p{2.2cm} X >{\centering\arraybackslash}p{2.2cm}}
\toprule
\textbf{Original Text} &\textbf{Corruption Type} & \textbf{Corrupted Text} & \textbf{Similarity (Text / Vision)} \\
\midrule
The morning sun filtered through the trees, casting golden patterns on the ground. & Character-level & T\textcolor{red}{eh} morn\textcolor{red}{n}ig sun f\textcolor{blue}{ltier}d \textcolor{red}{trough} the \textcolor{blue}{teers}, \textcolor{red}{sa}ting go\textcolor{red}{dl}en patt\textcolor{red}{re}ns on the gr\textcolor{green}{0}\textcolor{red}{nu}d. & 0.53 / \textbf{0.90} \\
\midrule
She sipped her coffee slowly, savoring the rich aroma and warmth. & Character-level & She si\textcolor{blue}{p}ed her \textcolor{blue}{cof}fee slo\textcolor{blue}{o}wly, sav\textcolor{green}{0}ring the r\textcolor{green}{!}ch aro\textcolor{red}{am} and warmth. & 0.68 / \textbf{0.92} \\
\midrule
The morning sun filtered through the trees, casting golden patterns on the ground. & Word-level & The \textcolor{blue}{\sout{morning}} \textcolor{green}{sunlight} \textcolor{green}{filter} through the \textcolor{blue}{\sout{trees, casting golden patterns}} \textcolor{red}{the on} \textcolor{green}{ground}. & 0.69 / \textbf{0.81}\\
\midrule
She sipped her coffee slowly, savoring the rich aroma and warmth. & Word-level & She \textcolor{blue}{\sout{sipped}} her \textcolor{green}{java} slowly, the rich \textcolor{green}{people} \textcolor{red}{warmth and}. & 0.61 / \textbf{0.86} \\
\midrule
Human mind does not read every letter by itself, but the word as a whole.& Typoglycemia & Hu\textcolor{red}{am}n m\textcolor{red}{ni}d d\textcolor{red}{eo}s not r\textcolor{red}{ae}d e\textcolor{red}{rve}y l\textcolor{red}{tete}r by i\textcolor{red}{stle}f, but the w\textcolor{red}{ro}d as a w\textcolor{red}{loh}e. & 0.60 / \textbf{0.88}\\
\midrule
According to a research team at Cambridge University, it doesn't matter in what order the letters in a word are, the only important thing is that the first and last letter be in the right place. & Typoglycemia & A\textcolor{red}{occdrni}g to a r\textcolor{red}{scheearc}h at C\textcolor{red}{mabrigd}e U\textcolor{red}{inervtis}y, it d\textcolor{red}{eosn'}t m\textcolor{red}{ttae}r in w\textcolor{red}{ah}t or\textcolor{red}{ed}r the l\textcolor{red}{ttee}rs in a w\textcolor{red}{ro}d are, the o\textcolor{red}{ln}y i\textcolor{red}{prmoet}nt t\textcolor{red}{ih}ng is t\textcolor{red}{ah}t the f\textcolor{red}{ri}st and l\textcolor{red}{sa}t l\textcolor{red}{tte}er be at the r\textcolor{red}{ghi}t p\textcolor{red}{cla}e. & 0.71 / \textbf{0.88} \\
\bottomrule
\end{tabularx}
\label{tab:text_corruption_show}
\end{table*}
\begin{table}[t]
\centering
\vspace{-0.2cm}
\caption{Zero-shot accuracy on the TKEval Character Count dataset~\cite{chai2024tokenization}. Results show that vision-centric inputs effectively mitigate the ``character blindness" of traditional subword tokenization.}
\label{tab:char_counting}
\resizebox{0.9\columnwidth}{!}{
\renewcommand{\arraystretch}{1.2} % 增加一点行高更美观
\begin{tabular}{llc}
\toprule
\textbf{Model} & \textbf{Input Type} & \textbf{Accuracy (\%)} \\
\midrule
Qwen2.5-VL 3B & Pure-Text & 57.99 \\
Qwen2.5-VL 3B & Visual-Text & 63.83 \greenp{5.84$\uparrow$}  \\
 \our &  Visual-Text & \textbf{64.98} \greenp{6.99$\uparrow$} \\
\bottomrule
\end{tabular}
}
\vspace{-0.3cm}
\end{table}

\subsubsection{Fertility}
We analyze the Fertility (FET), defined as the average number of tokens required to represent a single word. 
As summarized in Table~\ref{tab:high} and Figure~\ref{fig:lowsource} (left), \our\ consistently yields significantly lower fertility than standard text tokenization across all languages. 
Specifically, for high-resource languages, \our\ achieves an average FET of 0.37, a sharp contrast to the 2.21 FET of the text-tokenization baseline. 
This trend is even more pronounced in low-resource settings, where \our\ maintains a stable FET of 0.48, while text tokenization escalates to 3.88.
These results underscore a critical tokenization bias in traditional subword-based models: while they favor high-resource languages (\eg, English), they excessively fragment low-resource scripts—in extreme cases like Georgian (ka), words are decomposed to the character level with a FET of 8.33. 
In contrast, \our\ treats text as raw visual signals, ensuring cross-lingual equity by providing a compact and uniform representation regardless of the script's rarity. 
To provide a more granular perspective, we present a comprehensive language-wise breakdown of these compression ratios in Figure~\ref{fig:compression_ratio}. 
This ratio is calculated on the Flores-200 benchmark~\cite{costa2022no} by dividing the sequence length generated by the standard tokenizer (Qwen2.5-VL 3B~\cite{qwen}) by that of our vision-centric approach. 
As illustrated, \our\ achieves a consistent average compression of $5.71\times$ for high-resource languages, which increases to $7.85\times$ in low-resource settings. 
The consistent gains across the entire linguistic spectrum further validate \our\ as a highly efficient and scalable alternative to discrete text-tokenization.

% \subsection{Fine-Grained Lexical Reasoning}
% \label{sec:4.4}

\subsection{Perturbation Probing} 
\label{sec:perturbation}
We assess the robustness of Qwen2.5-VL 3B with text tokenization \vs\ \our\ with vision tokenization on MMLU~\cite{hendrycksmeasuring} under three perturbation types in a \textbf{zero-shot setting} (\ie, without any dataset-specific fine-tuning).
(i) \emph{Character-level noise.} 
For low-level surface corruption, we use the TKEval-MMLU~\cite{chai2024tokenization}, which simulates realistic typographical errors by applying within-word $n$-gram shuffling ($n \in \{2,3,5\}$) and random character edits such as insertions and deletions.
(ii) \emph{Visual attacks.} To evaluate perceptual robustness, we follow ECES~\cite{eger2019text}, substituting Latin letters with visually similar glyphs (\eg, ê for ``e”) at controlled perturbation levels $p \in \{0.2, 0.4, 0.6\}$.
(iii) \emph{Word-level noise.} To probe semantic robustness, words are randomly corrupted with probabilities $p \in \{0.2, 0.4, 0.6\}$, including synonym substitution and deletion.
More perturbation implementation details can be found in the Sec.~\ref {sec:pertubation_imple}.
\begin{table}[t]
\centering
\vspace{-0.2cm}
\caption{Zero-shot accuracy on the TKEval Word Unscrambling dataset (3,200 samples)~\cite{chai2024tokenization}. Results demonstrate that \textbf{vision-centric tokenization} enhances the ability of the model to reconstruct words from disordered character sequences.}
\label{tab:word_unscrambling}
\resizebox{0.9\columnwidth}{!}{
\renewcommand{\arraystretch}{1.2}
\begin{tabular}{llc}
\toprule
\textbf{Model} & \textbf{Input Type} & \textbf{Accuracy (\%)} \\
\midrule

Qwen2.5-VL 3B & Pure-Text & 10.94 \\
Qwen2.5-VL 3B & Visual-Text & 11.87 \greenp{0.93$\uparrow$} \\

 \our & Visual-Text & \textbf{12.50} \greenp{1.56$\uparrow$}\\
\bottomrule
\end{tabular}

}
\vspace{-0.3cm}
\end{table}

\subsubsection{Perturbation Implementation Details}
\label{sec:pertubation_imple}
In this paper, we consider three types of perturbations: character-level, word-level, and visual attacks. 
Examples are shown in the Table~\ref{tab:text_corruption_show}.
Across both word- and character-level perturbations, the similarity scores obtained using vision-centric tokenization consistently outperform those from text tokenization, demonstrating the superior robustness of our vision-centric approach.
Table~\ref{tab:app:visualnoise} showcases more examples of visual attack.

\noindent\textbf{Character-level Perturbation.} 
Following~\cite{chai2024tokenization}, we shuffle characters within word boundaries using $n$-grams of sizes 2, 3, and 5 with a probability of 50\%. We also apply $n$-gram noise by randomly inserting, deleting, or replacing characters, spaces, and punctuation marks to simulate spelling noise.
This corruption occurs with a probability of 30\%. 

\noindent\textbf{Word-level Perturbation.} 
Words are randomly perturbed with probabilities $p \in \{0.2, 0.4, 0.6\}$ through synonym substitution, internal word reordering, and deletions. 

\noindent\textbf{Visual Attack.} 
In line with~\cite{eger2019text}, each of the 26 uppercase and lowercase letters is substituted with a visually similar letter at varying probabilities $p \in \{0.2, 0.4, 0.6\}$.

\begin{figure*}[!t]
    \centering
    \includegraphics[width=1.\textwidth]{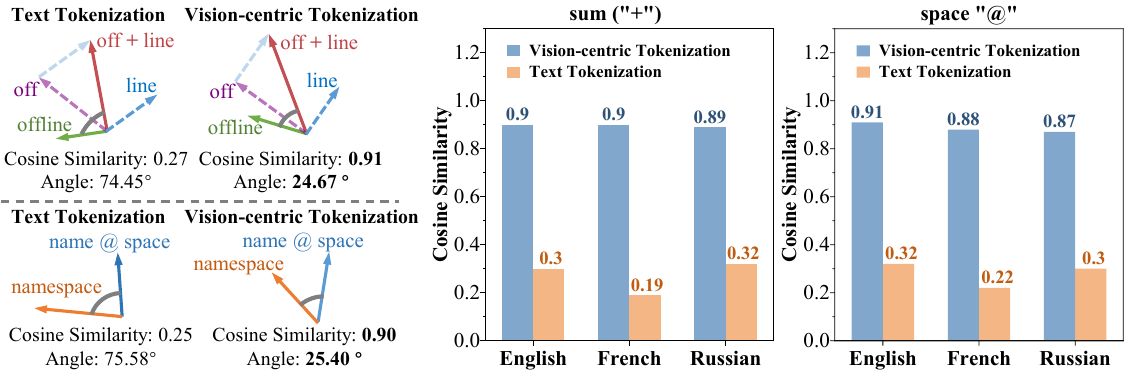}

\caption{\small{
\textbf{Compositional evaluation of token embeddings} from text and vision tokenization across three languages.
Cosine similarity and angle are computed between original full-word embedding (\eg, \texttt{offline}) and its composed embedding (\eg, \{\texttt{off}, \texttt{line}\}).
\textbf{Sum (``+”)} means the composed embedding is obtained by summing subword embeddings. 
\textbf{Space (``@”)} denotes composition by concatenating subwords with a space. 
\textbf{Vision tokenization yields composed embeddings more consistent with the full word across all languages.}
}
}
\label{fig:composition}
\vspace{-0.4cm}
\end{figure*}

\begin{table}[t]
\centering
\caption{\small \our\ consistently improves instruction-following with visual-text inputs across different model backbones. ${\dagger}$ corresponds to \our\ on JanusPro 1B~\cite{chen2025janus}, ${\ddagger}$ on Qwen2.5-VL 7B~\cite{qwen}, and ${\diamond}$ on Llava-next 8B~\cite{liu2024llavanext}.}
\label{tab:morellm}
\resizebox{0.85\linewidth}{!}{
\renewcommand\arraystretch{1.2}
\begin{tabular}{ccc}
\toprule
\textbf{Models} & \textbf{Text Source} & \textbf{TriviaQA} \\
\midrule
JanusPro 1B & Pure-Text & 42.71 \\
JanusPro 1B & Visual-Text & 27.10 \\
+ \our$^{\dagger}$ & Visual-Text & \textbf{35.23\greenp{8.13$\uparrow$}} \\
\hdashline
Qwen2.5-VL 7B & Pure-Text & 58.90 \\
Qwen2.5-VL 7B & Visual-Text & 53.53 \\
+ \our$^{\ddagger}$ & Visual-Text & \textbf{59.65\greenp{6.12$\uparrow$}} \\
\hdashline
Llava-next 8B & Pure-Text & 58.71 \\
Llava-next 8B & Visual-Text & 51.18 \\
+ \our$^{\diamond}$ & Visual-Text & \textbf{59.72\greenp{8.54$\uparrow$}} \\
\bottomrule
\end{tabular}}
\vspace{-0.4cm}
\end{table}

\subsubsection{Results}
Figure~\ref{fig:noise} presents the performance degradation on the MMLU benchmark under three types of orthographic perturbations: character-level, visual, and word-level noise. 
Across all noise intensities and categories, the vision-centric tokenization (\textbf{Visual-Text Input}) consistently exhibits superior robustness, suffering markedly smaller accuracy drops compared to the traditional subword-based (\textbf{Pure-Text Input}) counterpart. Specifically, as the perturbation level $p$ or the $n$-gram granularity increases, the performance gap between the two methods widens, highlighting the inherent fragility of text-based tokenizers. 
We attribute this to the fact that subword tokenization is highly sensitive to character-level fluctuations; even a minor typo can trigger a complete shift in the resulting token sequence, leading to semantic misalignment. 
In contrast, the vision tokenization treats characters as visual units, capturing their shape and spatial layout. 
Minor typographical or lexical changes affect only local visual details, leaving the overall representation largely intact and improving robustness to noise.

\begin{table}[!t] % 适配双栏排版，置于单栏顶部
  \centering
  \small
  % 移除了 width=0.4\textwidth 的限制，确保标题在单栏内自然换行
  \captionsetup{font=small}
  \setlength\tabcolsep{6pt} % 稍微增加列间距，避免内容太挤
  
  \caption{\small \textbf{Ablation on fine-tuning scope.} Keeping the projector frozen is critical for stable gains, with tuning the vision encoder and LLM providing optimal performance.}
  \label{tab:lora}
  
  % \vspace{-5pt} % 标题与表格之间的间距
  
  % 将宽度设为 1.0\columnwidth，这是单栏排版的标准宽度
  \resizebox{0.9\columnwidth}{!}{
    \renewcommand\arraystretch{1.2}
   \begin{tabular}{cccc}
\thickhline
\textbf{Vision Encoder} & \textbf{Projector} & \textbf{LLM} &\textbf{TriviaQA}  \\

\hline
& & & 37.55 \\
\cmark& & &40.12\\
& \cmark& & 31.93\\
& & \cmark&34.16\\
& \cmark& \cmark& 32.92\\
\cmark & \cmark &  \cmark & 37.02 \\
\cmark & & \cmark &\textbf{43.53} \\
\bottomrule
\end{tabular}
  }
  \vspace{-0.4cm}
  % \vspace{-10pt} % 调整表格与下方正文的间距
\end{table}

\subsection{Probing Intra-token Structural Fidelity}
\label{sec:intratoken}
Standard subword tokenization~\cite{kudo2018subword} converts text into sequences of discrete token IDs, inherently discarding the internal orthographic structure of words. 
As a result, models may struggle to perceive how words are composed, leading to degraded performance on tasks that require fine-grained lexical perception.
To investigate whether our vision-centric approach can effectively mitigate these tokenization-related constraints, we evaluate three diagnostic probes: \textbf{(i) subword compositionality, (ii) character counting, and (iii) word unscrambling~\cite{srivastava2023beyond}}. 
These experiments evaluate the sensitivity of the model to intra-token morphological structures, which are typically obscured in discrete symbolic mappings but natively preserved through vision-based representations.

\begin{table*}[t!]
%\vspace{0.1cm}
\captionsetup{font=small}
\caption{ \small{Evaluation under different training and inference text source settings. $^{*}$ indicates results on a reduced training dataset (145k samples), where long samples were removed to prevent out-of-memory issues with pure-text input.
${^\clubsuit}$ denotes the same finetuning setup as \our, with pure-text training input.
}
}
% \vspace{-0.2cm}
\centering
\small
\resizebox{1.\textwidth}{!}{
\renewcommand\arraystretch{1.4}
\begin{tabular}{cccccccc}
\thickhline

\textbf{Models}  & \textbf{Training Input} & \textbf{Inference Input} & {\textbf{TriviaQA}}	&{\textbf{NQ}}&	{\textbf{PopQA}}&	{\textbf{SST5}} & \textbf{MMLU}   \\
\hline 

 Qwen2.5-VL 3B & - & Visual-Text & {37.55}& {21.13}& {20.16}&  {25.21}& 32.31	 \\
 Qwen2.5-VL 3B & - & Pure-Text & {41.92}& {29.31} &{24.64}&{28.80}& 61.91	 \\

% + \our & Visual-Text & Pure-Text & \textbf{43.53\greenp{5.98$\uparrow$}}&	24.14\textbf{\greenp{3.01$\uparrow$}}&	24.26\textbf{\greenp{4.10$\uparrow$}}
%  &\textbf{44.40\greenp{19.19$\uparrow$}}&
%  52.52\textbf{\greenp{20.21$\uparrow$}} \\
\arrayrulecolor{gray}\hdashline\arrayrulecolor{black}

Qwen2.5-VL 3B$^{\clubsuit}$ & Pure-Text & Pure-Text &  
{42.06\greenp{0.14$\uparrow$}}	&
{29.75\greenp{0.44$\uparrow$}}&	
{24.96\greenp{0.32$\uparrow$}}&
{30.00\greenp{1.20$\uparrow$}} &
62.21\textbf{\greenp{0.30$\uparrow$}}    \\

+ \our$^{*}$ & Visual-Text & Visual-Text & 
{42.18\textbf{\greenp{4.63$\uparrow$}}}&
{23.16\textbf{\greenp{2.03$\uparrow$}}}&	
{23.47\textbf{\greenp{2.03$\uparrow$}}}&
{32.80\textbf{\greenp{7.59$\uparrow$}}}&
49.00\textbf{\greenp{16.70$\uparrow$}}    \\

+ \our$^{*}$ & Visual-Text & Pure-Text & 
{42.54\greenp{0.62$\uparrow$}}&	
{30.18\greenp{0.62$\uparrow$}}&	
{25.21\greenp{0.62$\uparrow$}}&	
{31.42\greenp{2.62$\uparrow$}}&
62.34\greenp{0.43$\uparrow$}

% \textbf{43.53\greenp{5.98$\uparrow$}}&	24.14\textbf{\greenp{3.01$\uparrow$}}&	24.26\textbf{\greenp{4.10$\uparrow$}}
%  &\textbf{44.40\greenp{19.19$\uparrow$}}&
%  52.52\textbf{\greenp{20.21$\uparrow$}} 
\\
\bottomrule

\end{tabular}}
\vspace{-10pt}
\label{tab:discussion}
\end{table*}
\begin{table}[!t] % 去掉星号，改为单栏 [!t] 建议放在页首
    \centering
    \captionsetup{font=small}
    \caption{\small Impact of data scaling on \our. Results demonstrate a monotonic performance gain across all reasoning and QA tasks as the volume of visual-text instruction data increases.}
    \label{tab:app:scaling}
    
    % 使用 \columnwidth 确保表格宽度正好契合单栏
    % 如果表格内容不多，可以去掉 \resizebox 直接用 \small
    \resizebox{\columnwidth}{!}{
        \renewcommand\arraystretch{1.1} % 略微增加行高，提升阅读感
        \setlength\tabcolsep{6pt}      % 微调列间距，防止单栏下太拥挤
        
        \begin{tabular}{c c c c c} % 第一列左对齐，数值列居中
            \toprule
            \textbf{Training Size} & \textbf{TriviaQA} & \textbf{NQ} & \textbf{MMLU} & \textbf{SST-5} \\
            \midrule
            0k    & 37.55 & 21.13 & 32.31 & 25.21 \\
            9k    & 40.27 & 22.31 & 41.22 & 30.60 \\
            145k  & 42.18 & 23.16 & 49.00 & 32.80 \\
            658k  & \textbf{43.53} & \textbf{24.14} & \textbf{52.52} & \textbf{44.40} \\
            \bottomrule
        \end{tabular}
    }
\vspace{-0.3cm}
\end{table}

\subsubsection{Subword Compositionality} 
\label{subwordcom}
\noindent\textbf{Task Description and Settings.}
Compositional ability enables the model to \emph{generalize to novel combinations} instead of just memorizing patterns~\cite{chai2024tokenization,peng2025understanding}.
To assess the ability of text- and vision-tokenized embeddings to capture subword compositional structure, we draw on the SIGMORPHON 2022 dataset~\cite{batsuren2022sigmorphon}, which provides full words and their possible subword decompositions (\eg, \texttt{offline} $\rightarrow$ \texttt{off, line}).
As in~\cite{peng2025understanding}, we retain only full words that appear in the model vocabulary and perform experiments across English, French, and Russian.
We measure cosine similarity and angle between full-word embedding and its corresponding composed embedding to evaluate compositional fidelity.
The composed embeddings are constructed in two ways: (i) \emph{sum}, by summing the embeddings of each subword, and (ii) \emph{space}, by embedding the subwords concatenated with a space.

\noindent\textbf{Results.} 
Figure~\ref{fig:composition} shows vision tokenization achieves cosine similarity close to 1.0 and much smaller angles than text tokenization across all languages.
This suggests that \textit{vision-based embeddings capture compositional structure far more faithfully}, as they encode each word as a sequence of visual patterns, inherently maintaining local geometric relations.
By contrast, the text tokenization splits the word into independent subword units without explicitly modeling the hierarchy from characters to words.
This limitation not only weakens compositional alignment but also \emph{explains the greater sensitivity of text-tokenized embeddings to surface-level perturbations and morphological changes}.

\subsubsection{Character Counting}
\label{sec:ccount}
\noindent\textbf{Task Description and Settings.}
We evaluate the ability of the model to resolve fine-grained textual details through a zero-shot character counting task on the TKEval character count dataset~\cite{chai2024tokenization}. 
This task requires the model to identify the frequency of a target letter within a given string—a notorious challenge for subword-based LLMs (\eg, How many times does the letter r appear in `strawberry'?) because the internal character structure is collapsed into atomic token IDs. 
We randomly sample 5,000 instances from the dataset~\cite{chai2024tokenization} and measure zero-shot accuracy.

\noindent\textbf{Results.} 
Table~\ref{tab:char_counting} illustrates the performance gap between text and vision-centric tokenization. 
To ensure a controlled comparison, we evaluate a frozen Qwen2.5-VL 3B baseline~\cite{bai2025qwen2}. 
Transitioning from text-based tokenization to vision-centric tokenization yields a significant accuracy boost of +5.84\% (from 57.99\% to 63.83\%), confirming that representing text via visual patterns effectively recovers the orthographic information lost in subword-based IDs. Furthermore, \our\ leverages this vision-centric paradigm to achieve a peak accuracy of 64.98\%, demonstrating superior robustness in fine-grained structural reasoning.
Critically, $\our$ achieves this superior performance in a strict zero-shot setting, having never been exposed to character-counting tasks or related datasets during training.

\begin{table}[t] % 去掉星号，改为单栏显示；建议使用 [t] 放在页首
\centering
\small % 略微缩小字体以适配单栏宽度，若空间充裕可删掉
\caption{\textbf{Robustness to Font Types.} Although trained on Noto Sans, \our\ maintains stable performance on unseen font types (\eg, Arial, Georgia).}
\label{tab:font_robustness} % 建议加上 label 方便正文引用
\resizebox{0.9\columnwidth}{!}{
\renewcommand{\arraystretch}{1.2}
\begin{tabular}{c c c c} % 建议首列左对齐 (l)，数值列居中 (c)

\toprule % 顶线
\textbf{Font Type} & \textbf{TriviaQA} & \textbf{MMLU} & \textbf{SST5}\\
\midrule % 中线
Noto Sans (Train) & 43.53 & 52.52 & \textbf{44.40} \\
Arial             & 43.47 & \textbf{52.87} & 43.97 \\
Georgia           & \textbf{43.62} & 52.36  & 43.99\\
\bottomrule % 底线
\end{tabular}
}
\label{tab:font_style}
\end{table}

\begin{table}[!t] % 使用标准 table，[!t] 置于栏顶部
  \centering
  \small
  % 移除了 width=0.4\textwidth 限制，让标题占满单栏
  \captionsetup{font=small} 
  \setlength\tabcolsep{6pt} % 稍微增加了一点间距（原2pt在单栏下可能太挤）
  \vspace{0.3cm}
  \caption{\small
   \our\ matches Qwen2.5-VL 3B on VQAv2, DocVQA, and TextVQA, showing that vision-centric instruction tuning preserves naive vision–language performance.
  }
  \label{tab:vqa}
  
  % \vspace{-5pt} % 标题与表格间的标准间距
  
  % 适配单栏宽度
  \resizebox{0.9\columnwidth}{!}{
    \renewcommand\arraystretch{1.4}
    \begin{tabular}{cccc}
      \thickhline
      \textbf{Model} & {\textbf{VQAv2}} & {\textbf{DocVQA}} & {\textbf{TextVQA}} \\
      \hline
      {Qwen2.5-VL 3B} & {81.2} & {93.9} & {79.3} \\
      {\our} & {81.4} & {93.5} & {80.1} \\
      \bottomrule
    \end{tabular}
  }
  
  \vspace{-10pt} % 调整表格下方与正文的空白
\end{table}

\subsubsection{Word Unscrambling}
\label{sec:wordunscram}
\noindent\textbf{Task Description and Settings.}
To investigate the capacity of the model for fine-grained structural manipulation, we conduct \textit{zero-shot} experiments on the Word Unscrambling task using the TKEval-WU dataset (3,200 samples)~\cite{chai2024tokenization}. 
The task requires the model to recover the original word from a randomly scrambled sequence of characters (\eg, \texttt{"nad"} $\rightarrow$ \texttt{"and"}). 
We evaluate the zero-shot accuracy to assess if the vision-centric approach can better capture the character-level nuances indispensable for text reassembly.

\noindent\textbf{Results.} 
The results are summarized in Table~\ref{tab:word_unscrambling}. 
The low overall accuracy across all models underscores the inherent difficulty of the task. 
However, the paradigm shift from text-based to vision-centric tokenization yields consistent improvements. 
In the controlled comparison using a frozen Qwen2.5-VL-3B, vision-centric tokenization-based model outperforms text tokenization-based baseline (+0.93\%), suggesting that visual patterns provide a more stable foundation for character-level reasoning than discrete IDs. 
\our\ achieves the best performance with 12.50\% accuracy \textit{despite not being explicitly trained on the word unscrambling task}. 
These findings confirm that by maintaining the spatial and morphological integrity of text, vision-centric tokenization allows the model to resolve better and reorder internal constituents, even when the input surface form is heavily distorted.

% \begin{wrapfigure}{r}{0.3\textwidth} 

%     \centering
%     \includegraphics[width=0.3\textwidth]{figs/src/layer.pdf} 

%     \caption{\small{Layer-wise residual norms from orthogonal Procrustes alignment between visual-text and pure-text embeddings. Instruction tuning lowers residual norm in deeper layers, reflecting more consistent processing of pure-text and visual-text inputs.}}
%     \label{fig:layer}
% \end{wrapfigure}
% % ~\citep{schonemann1966generalized}

\begin{figure}[t]
\centering
\vspace{-0.2cm}
\includegraphics[width=0.75\linewidth]{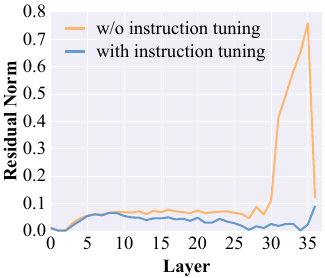}
\caption{\small Layer-wise residual norms from orthogonal Procrustes alignment between visual-text and pure-text embeddings. Vision-centric instruction tuning lowers residual norm in deeper layers (31--35), reflecting more consistent processing of pure-text and visual-text inputs.}
\label{fig:layer}
\vspace{-0.4cm}
\end{figure}
\subsection{Ablation Study}

\subsubsection{Extension to More MLLMs} 
To prove the generality of \our, we test the unified model JanusPro 1B~\cite{chen2025janus}, Qwen2.5-VL 7B~\cite{bai2025qwen2}, and Llava-next-8b~\cite{liu2024llavanext} on TriviaQA~\cite{joshi2017triviaqa} under both pure-text and visual-text inputs, comparing performance with and w/o \our. 
As summarized in Table~\ref{tab:morellm}, both backbones show degraded performance with visual-text inputs, as they have not been exposed to such instructions during pretraining. 
However, integrating \our\ yields substantial gains, recovering or even exceeding their performance on pure-text inputs.
These results confirm \our\ consistently improves instruction-following in visual-text settings.

\subsubsection{Scaling Study on Visual-Text Instruction Data}
\label{app:scaling}
To evaluate the scalability of \our, we conduct a controlled study by training the model with varying volumes of visual-text instruction data—specifically 9k, 145k, and 658k examples—sampled from OpenHermes 2.5~\cite{OpenHermes2.5}. 
As illustrated in Table~\ref{tab:app:scaling}, performance across all benchmarks exhibits a consistent improvement as the data scale increases. 
Notably, the accuracy on the reasoning-intensive MMLU benchmark surges from 32.31\% to 52.52\%, representing a significant 20.21\% absolute gain. 
These results underscore the potential of our vision-centric paradigm to further enhance complex reasoning capabilities through continued data expansion.

\subsubsection{Effect of Font Type} 
\our\ is finetuned with the \textit{Google Noto Sans} font. 
At inference time, we render the same textual content using two additional, \textbf{unseen font families} (Arial, Georgia) and evaluate performance under the same protocol. 
As illustrated in Table~\ref{tab:font_style}, the performance remains comparable or even slightly improves, showing that the model is robust to variations in font style.

\subsubsection{Ablation on Fine-tuning Scope} 
To identify the most effective strategy for integrating LoRA into our multimodal framework, we conduct a comprehensive ablation study across three primary components: the vision encoder, the projector, and the Large Language Model (LLM).
As summarized in Table~\ref{tab:lora}, when all components are frozen (baseline), the model achieves a score of 37.55. 
We observe that solely adapting the vision encoder yields a noticeable improvement (+2.57), whereas fine-tuning the LLM or the projector in isolation leads to a performance drop. 
Most notably, the configuration that simultaneously adapts the vision encoder and the LLM while keeping the projector frozen achieves the superior result of 43.53, surpassing the baseline by 5.98 points.
We attribute this phenomenon to two factors. 
First, adapting the vision encoder allows the model to extract more task-relevant visual features, while tuning the LLM enables better instruction following. Second, the projector, which is pre-trained on massive image-text pairs, has already established a robust and generalized cross-modal alignment. 
Fine-tuning it on the relatively limited instruction-tuning data likely causes alignment disruption, leading to catastrophic forgetting of the pre-learned multimodal mapping.
Therefore, maintaining a frozen projector is essential for preserving stable and optimal performance gains.

\subsubsection{Preservation of Vision-native Capabilities}  
To verify whether vision-centric instruction tuning compromises the fundamental multimodal proficiencies of the MLLM, we evaluate \our\ on several standard vision-native benchmarks, including VQAv2~\cite{goyal2017making}, DocVQA~\cite{mathew2021docvqa}, and TextVQA~\cite{singh2019towards}.
As summarized in Table~\ref{tab:vqa}, \our\ achieves highly competitive performance that is on par with the Qwen2.5-VL 3B baseline across all tasks. 
Specifically, while maintaining nearly identical accuracy on VQAv2 and DocVQA, \our\ even demonstrates a slight improvement on TextVQA (80.1\% vs. 79.3\%). 
These results provide rigorous evidence that our vision-centric paradigm effectively preserves the original vision-language reasoning capabilities, ensuring that the enhanced instruction-following performance does not come at the cost of native task proficiency.

\section{Discussion}
\label{sec::discussion}

\subsection{Do Visual-Text Instruction Improvements Only Stem from Additional Knowledge from the New Data?}
A key question is whether the gains observed after visual-text instruction finetuning arise from access to new knowledge in the finetuning corpus, or from improved ability to follow visual-text instructions.  
To disentangle these factors, we finetune Qwen2.5-VL 3B on the same data in two formats: visual-text and pure-text.  
Because pure-text input consumes substantially more tokens, we further filter samples to avoid out-of-memory issues (denoted by * in Table~\ref{tab:discussion}, Rows 4–6).
The results reveal a striking contrast: visual-text finetuning (row 5) yields a \textbf{+16.70 improvement} over baseline (row 2), whereas pure-text finetuning offers only a marginal gain of +0.30.  
This indicates that the improvements \textit{primarily stem from enhanced instruction-following ability in the visual-text format}, rather than access to new information.
Moreover, the efficiency of visual-text tokens enables training on more examples under identical compute constraints, producing even larger gains (52.52 \vs\ 49.00). 
Thus, the advantage of our approach lies not only in robustness to tokenization but also in more effective use of limited training budgets.

\subsection{Enhanced Text-Only Performance after Visual-Text Instruction Tuning}
We examine how fine-tuning the model with visual-text instructions impacts its performance on pure-text inputs, using five widely recognized benchmarks for evaluation.
As detailed in Table~\ref{tab:discussion}, Qwen2.5-VL 3B~\cite{qwen} finetuned on visual-text input (Row 6) achieves greater improvements on pure-text inference than the variant finetuned on pure-text data (Row 4), consistently across the five benchmarks.
This improvement suggests that \textit{even though the finetuning is performed using visual-text data, the model benefits from better cross-format generalization, enhancing its pure text performance}. 
The ability to process both image-based and text-based instructions likely equips the model with richer understanding capabilities that extend beyond the specific input format.
Notably, finetuning with visual-text inputs is more efficient, as it uses far fewer input tokens than pure-text finetuning, allowing the model to achieve stronger improvements at lower computational cost.

\subsection{Layerwise Effect of Instruction Tuning on Cross-Modal Alignment}
A key question is whether instruction tuning helps the model treat visual-text inputs consistently with their pure-text counterparts.
To probe this, we apply Orthogonal Procrustes analysis~\cite{schonemann1966generalized} on Qwen2.5-VL 3B, with 1k out-of-distribution samples from ALPAGASUS~\cite{chenalpagasus}.
This method finds the optimal linear transformation that aligns visual-text embeddings with pure-text embeddings while preserving internal geometry.
We quantify alignment using the residual norm, \ie, the Frobenius distance between the transformed visual-text embeddings and the corresponding pure-text embeddings. 
Lower residual norm indicates stronger structural similarity.
Results in Figure~\ref{fig:layer} reveal that instruction-tuned models achieve progressively lower residuals in deeper layers, reflecting improved convergence between text and visual-text pathways. 
In contrast, the frozen model exhibits high residuals in the layers 31–35, consistent with its weaker performance on visual-text instructions. 
These results suggest that instruction tuning reshapes representational geometry across modalities, enabling more consistent processing of pure-text and visual-text inputs.

\section{Conclusion}
In this work, we introduce \our, a simple yet effective vision-centric tokenization method that substitutes conventional text tokenization by encoding rendered images through pretrained vision encoders.
Our approach achieves competitive or superior performance to conventional text tokenization, while offering clear advantages in multilingual efficiency, compositionality, and robustness to noise.
These results highlight the promise of visual tokenization as a general alternative to prevailing subword tokenization.
In future work, we plan to leverage vision encoders as a unifying interface across modalities, paving the way toward more general multimodal reasoning.

\bibliographystyle{IEEEtran}
\bibliography{main}

\end{document}